\documentclass[a4paper]{article}
\usepackage{subcaption}
\usepackage{multirow}
\usepackage{INTERSPEECH2020}
\usepackage{cite}
\usepackage{color}
\usepackage{sectsty}
\usepackage{url}
\usepackage{tipa,vowel}
\usepackage[toc,page,header]{appendix}

\title{Filler Word Detection and Classification: A Dataset and Benchmark}
\name{Ge Zhu$^{1*}$\thanks{*This work was performed during an internship at Adobe Research.}, Juan-Pablo Caceres$^2$, Justin Salamon$^2$}
\address{$^1$University of Rochester, $^2$Adobe Research}
\email{ge.zhu@rochester.edu \text{} \{caceres,salamon\}@adobe.com} 
\begin{document}

\maketitle

\begin{abstract}
Filler words such as `uh' or `um' are sounds or words people use to signal they are pausing to think. Finding and removing filler words from 
recordings is a common and tedious task in media editing. Automatically detecting and classifying filler words could greatly aid in this task, but few studies have been published on this problem to date. A key reason is the absence of a dataset with annotated filler words for model training and evaluation. In this work, we present a novel speech dataset, PodcastFillers, with 35K annotated filler words and 50K annotations of other sounds that commonly occur in podcasts such as breaths, laughter, and word repetitions. We propose a pipeline that leverages VAD and ASR to detect filler candidates and a classifier to distinguish between filler word types. We evaluate our proposed pipeline on PodcastFillers, compare to several baselines, and present a detailed ablation study. In particular, we evaluate the importance of using ASR and how it compares to a transcription-free approach resembling keyword spotting. We show that our pipeline obtains state-of-the-art results, and that leveraging ASR strongly outperforms a keyword spotting approach. We make PodcastFillers publicly available, in the hope that our work serves as a benchmark for future research.
\end{abstract}


\section{Introduction}
Speech disfluencies, such as filler words, stuttering, repetitions and corrections, are common in spontaneous speech \cite{das2019increase}. 
Of all disfluencies, filler words, especially `uh's and `um's, are the most common~\cite{womack2012disfluencies}. For content creators working on, e.g., podcasts or video interviews, manually finding and editing filler words in video and audio recordings requires significant time and effort. Automatically detecting filler words accurately has the potential to significantly speed up speech content creation workflows. Such a filler word detection system must be able to both localize filler words in time and classify them correctly.

Previous work has focused on detecting and removing speech disfluencies from text transcripts\cite{bach2019noisy,wang2018semi,jamshidlou2018,wang2020multi}, some also incorporating acoustic features\cite{zayats2019giving}. In some cases, the transcripts are produced via Automatic Speech Recognition (ASR)~\cite{ferguson2015disfluency,hassan2014segmentation,heeman16_interspeech}. 
In this scenario it is up to the ASR to transcribe the filler words, which requires training an ad-hoc ASR with filler words in its vocabulary. This is computationally intensive and challenging since ASR systems are often trained on spoken text corpora which do not contain any filler words, and thus cannot detect them reliably. Furthermore, adding a new filler word to the vocabulary would require re-training the ASR model.

More recently, several data driven methods have been proposed to detect speech disfluencies directly from audio in telephone conversations\cite{gupta2013paralinguistic,lea2021sep} and naturalistic recordings\cite{kaushik2015laughter,sheikh2021stutternet,kourkounakis2021fluentnet}. Sheikh et al.~proposed StutterNet~\cite{sheikh2021stutternet}, a time-delay neural network (TDNN) to classify repetitions, blocks, prolongations and interjections, the latter being another term for filler words. They used the UCLASS dataset~\cite{howell2009university}, which is designed for stutter classification. 
This poses some challenges, since the dataset was recorded in a controlled environment, and exclusively contains speech by people who stutter. This, along with the small size of the dataset ($\sim$4K sentences), means it is unclear whether the results would generalize to recordings of spontaneous speech more broadly.
To tackle this issue, 
Kourkounakis et al.~\cite{kourkounakis2021fluentnet} expanded the training data by creating a synthesized stutter dataset, LibriStutter, by inserting repetitions or interjections in between non-stuttered speech from a subset of LibriSpeech \cite{panayotov2015librispeech}. 
The interjections, however, were taken from UCLASS, presenting the same aforementioned challenge. Salamin et al.~\cite{salamin2013automatic} trained Hidden Markov Models to segment laughter, fillers, speech and silence from spontaneous speech on the SSPNet Vocalization Corpus, which is also a small-scale dataset. Lea et al.~\cite{lea2021sep} created a large speech disfluency dataset, SEP-28K, from podcasts with people who stutter, and used it to build a stutter detector. As before, there is a generalization challenge given the audio data are specific to people who stutter. Also, the dataset is annotated at the clip level, so it does not provide precise timestamps for filler words and cannot be used to evaluate detection accuracy at a fine temporal resolution.


The closest study to our work is by Das et al.~\cite{das2019increase}, who proposed a disfluency repair system aiming at removing filler words and long pauses. They trained a convolutional recurrent neural network (CRNN) for filler word segmentation (detection and classification) applied directly to audio recordings. They used two speech datasets, Switchboard speech data~\cite{godfrey1992switchboard} with transcripts and Automanner \cite{tanveer2016automanner}. A key limitation of the approach noted by the authors is that it was unable to distinguish filler words such as `uh' or `um' from real parts-of-speech, returning false-positives for actual words that sound similar to or contain filler words, such as ``um-brella''. Also, Kaushik et al.~\cite{kaushik2015laughter} found that the mismatch between training on telephony speech and testing on naturalistic recordings hurts filler classification accuracy.
The test set for evaluating the methods presented by Das et al.~\cite{das2019increase} only contains 20 speech samples, once again making it hard to draw generalizable conclusions.

In this paper, we address the data scarcity challenge for filler word detection by creating the largest annotated dataset of filler words published to date, PodcastFillers, which we make publicly available online\footnote{\url{podcastfillers.github.io}}. We propose an efficient workflow for generating annotation candidates in continuous speech recordings that leverages a robust Voice Activity Detection (VAD) model and an off-the-shelf ASR, and annotate over 85K filler word candidates. The resulting dataset spans 145 hours of speech from over 350 speakers coming from 199 public podcast episodes, and has 35K annotated filler words and 50K annotations of other speech events that are common in podcasts such as laughter, breaths, and repetitions. It also includes the ASR transcriptions we obtained for all the episodes. Using PodcastFillers, we train a filler classifier similar to a keyword spotting approach, and present an end-to-end pipeline that leverages VAD, ASR, and the classifier to perform filler word detection and classification. We compare our proposed pipeline to two baselines and show that it yields state-of the-art results. We hope it serves as a robust benchmark for future research.

\section{The PodcastFillers Dataset}
\label{sec:data}
\subsection{Podcast curation} 
We manually curated 199 gender-balanced, English-language podcast episodes from SoundCloud\footnote{\url{www.soundcloud.com}}, totaling 145 hours of speech audio from over 350 speakers sampled at 44.1 kHz. We searched for episodes using 
the 
24 topic categories 
defined in~\cite{GigaSpeech2021} 
to include a variety of topics and styles, and selected episodes from different shows or shows with guest speakers to ensure a diversity of speakers.


\subsection{Filler word annotation pipeline} 
\label{sec:cp}
\begin{figure}[t]
\setlength{\belowcaptionskip}{-8pt}
\centering
\includegraphics[width=3.1in]{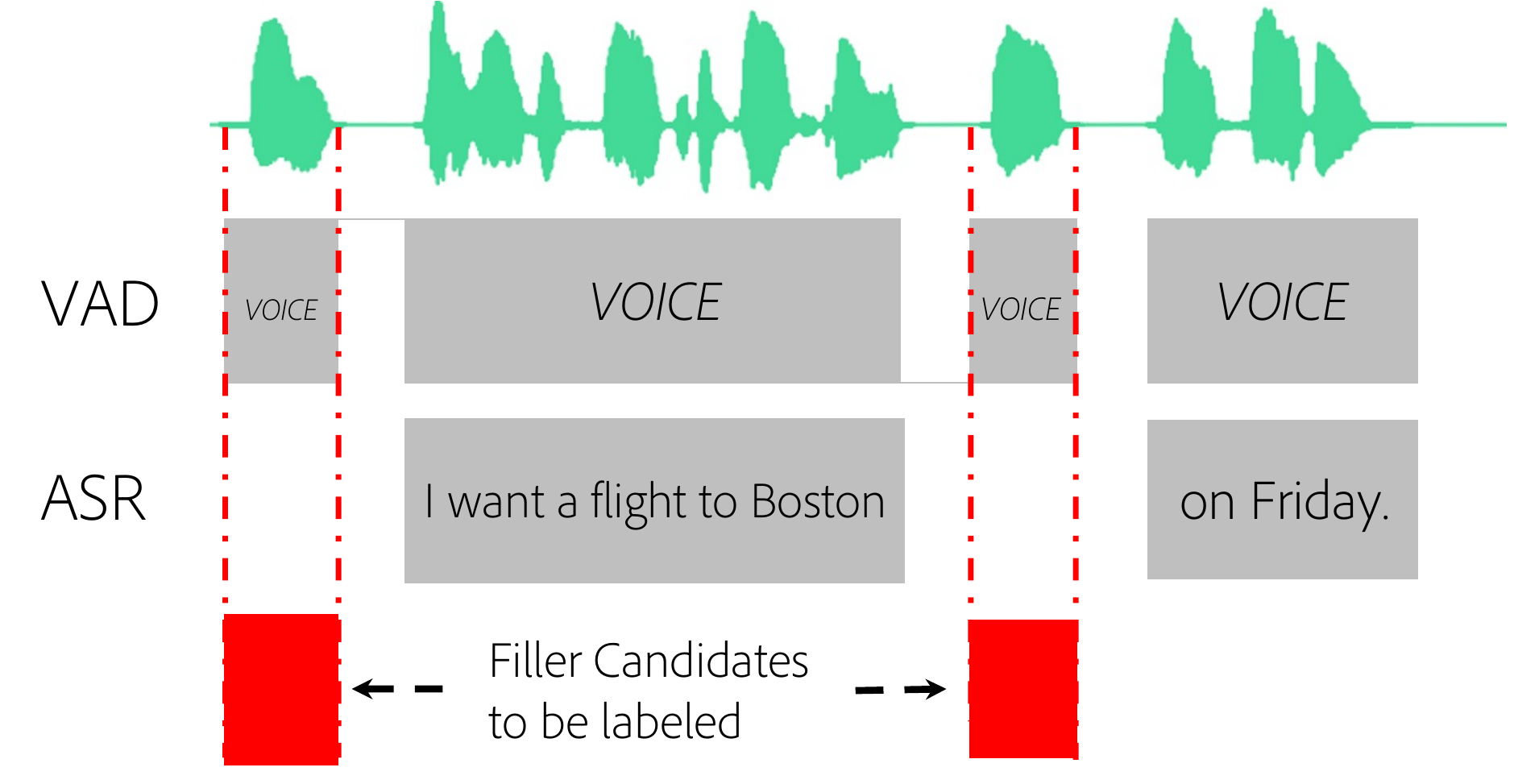}
\caption{Filler word candidate generation pipeline: non-linguistic filler word candidates are identified at times where VAD is activated while ASR is not.} 
\label{fig:filler_generation}
\end{figure}
Listening to the entire dataset to label fillers would be highly inefficient. Instead, we propose an annotation pipeline that leverages a commercial ASR system\footnote{\url{www.speechmatics.com}} and a VAD model to generate filler candidates for annotation, depicted in Fig.~\ref{fig:filler_generation}. 

As previously noted, ASR systems typically do not transcribe non-lexical fillers in spontaneous speech. We make use of this drawback to identify possible filler word locations: non-lexical fillers such as `uh' and `um' will trigger the VAD model, but appear as silent gaps in the ASR output. These regions where VAD activates but ASR does not are candidate locations for fillers. Using this approach we identified 85K candidates in PodcastFillers. Since the candidates may contain other sounds such as breaths, laughter, music, or even words (due to ASR errors), they require manual verification, which we obtained via crowdsourcing using a custom-built annotation interface.

\textbf{Voice Activity Detection (VAD) model and data:}
For detecting candidate fillers, we need a VAD model that outputs predictions at a fine temporal resolution (100 Hz) to precisely locate the temporal boundaries of speech regions. It also needs to be robust to various background and foreground noises in podcasts such as music and non-speech sounds (e.g., fan noise).

We achieve this fine temporal resolution by computing input acoustic features at a 10 ms hop size, such that we can slide the trained model over the audio at this temporal resolution. To ensure robustness, we need to combine a generalizable ML model with a varied training set containing various background and foreground noises at different signal-to-noise ratios (SNR). We create a new labeled speech dataset for VAD by programatically combining recordings of clean speech with music and noise using the Scaper soundscape mixing software \cite{salamon2017scaper}.

We generate frame-level (10 ms) VAD annotations by computing the audio amplitude from clean speech recordings sourced from the Librispeech-100~\cite{panayotov2015librispeech} and VCTK~\cite{veaux2016superseded} datasets, labeling regions below a 19 dB threshold relative to the peak amplitude of the normalized signal as silent. Then we programmatically mix the clean speech clips with background music and environmental sound from the strongly labeled subset~\cite{hershey2021benefit} of AudioSet ~\cite{gemmeke2017audio}. To test our VAD model, we keep a disjoint test set of audio source material using 8$\%$ of the speakers in VCTK, the test partition of Librispeech, and 5$\%$ of sound events from the AudioSet subset. We generate 300,000 training mixtures from the training source material and 10,000 test mixtures from the test source material.
Scaper allows us to control the SNR range and distribution in the mixtures. Preliminary experiments showed the performance of the VAD model, in terms of producing filler candidates when combined with ASR, was sensitive to the SNR range. We empirically settled on a speech SNR range of $[12,22]$ dB relative to background noise, $[-3, 17]$ dB relative to foreground noise and $[-6, 14]$ dB relative to music. 
We compute log-scaled mel-spectrograms (log-mel) as input to the VAD model using Librosa \cite{mcfee2015librosa}. We use 64 mel bins, and a purposely short window of 25ms and a hop size of 10ms, to support inference at a high temporal resolution. We adopt a Convolutional Recurrent Neural Netowrk (CRNN) architecture that has been shown to be robust for VAD in complex environments with noise~\cite{Chen2020voice}, but remove the recurrent layer to improve run-time performance. We found this change does not impact model accuracy. Our trained VAD model obtained Precision/Recall of 0.93/0.92 respectively on the test split of our mixed dataset.
%
Once our VAD model was trained, we used it in combination with the ASR to produce filler candidates. To minimize the chance of missing soft fillers, we set a lenient VAD activation threshold of 0.1 (as opposed to the standard 0.5 out of $[0,1]$). We found the majority of candidates to have a duration in the 150-400 ms range. For candidates shorter than 150 ms it was hard to determine by ear whether they were actual filler words or other sounds, so we decided to remove them prior to labeling. Similarly, we removed candidates longer than 2 s, which were rare and not representative of our target use case. 

\textbf{Labeling filler candidates:}
Based on an initial audition of a sample of candidates, we identified a set of filler and non-filler classes for our labeling task. 
%
%
The labels, along with the final number of annotations per label (in parentheses), are: For fillers, `uh' (17907), `um' (17078), `you know' (668), `like' (157), and `other' (315). `Like' and `you know' occurred rarely in our candidate set, when the ASR failed to transcribe them. For non-fillers, `laughter' (6623), `breath' (8288), `agreement sound' (3755, e.g., `mmm' or `uh-huh'), `regular words' (12709), `repetitions' (9024), `simultaneous speakers' (1484), `music' (5060) and `noise' (2735). The first three non-filler labels represent voice sounds that aren't fillers. The next three are caused by ASR errors or intentional omissions, and the final two are caused by VAD false-positives.
%

The candidates were presented to crowd workers for annotation. Each filler candidate was positioned at time 3 sec inside a 5 sec clip (for context), and highlighted in the interface. Annotators had to determine whether the highlighted candidate was a filler word or not, and based on that select one of the five filler labels or eight non-filler labels. Each  candidate was annotated by two people, or three when the first two disagreed.
Out of all candidates labeled `uh' or `um' 
in the dataset, 98.4\% and 96.4\% respectively had at least two annotators agree on the label.
%



\section{Filler Detection Pipeline}
\label{sec:method}

\begin{figure}[t]
\setlength{\belowcaptionskip}{-10pt}
\centering
\includegraphics[width=2.2in]{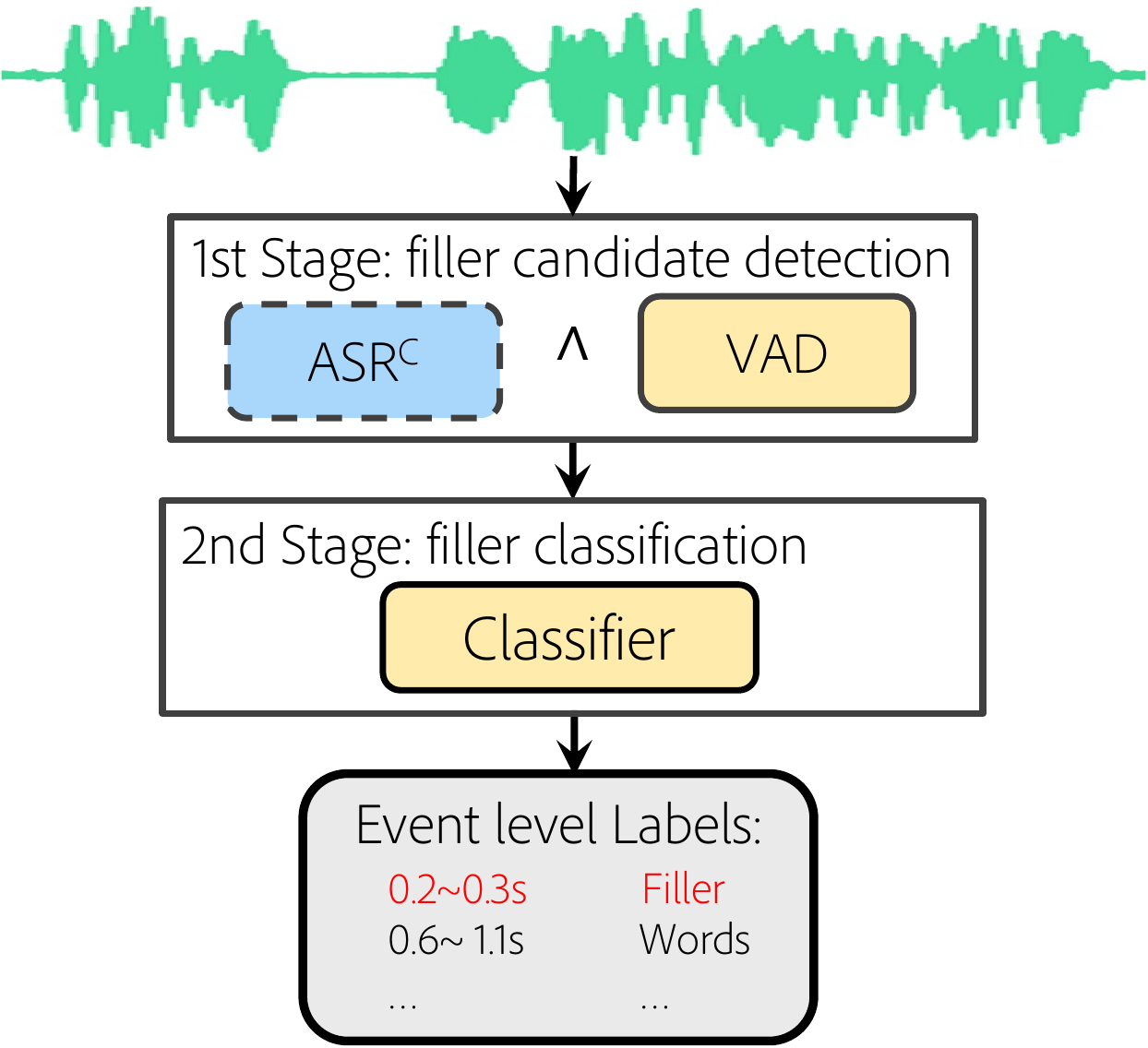}
\caption{Proposed two-stage filler detection and classification.}
\label{fig:system}
\end{figure}

We propose a filler detection pipeline with two variants: the first leverages ASR, while the second does not, which is relevant for deployment scenarios where ASR is not available.
The pipeline is depicted in Fig.~\ref{fig:system}.
In the first stage, the input audio is passed through the VAD model to find voice regions. The first pipeline variant also runs the audio through ASR to discard regions with transcribed words. The second variant passes straight to the next stage.
In the second stage, the remaining candidate time regions are passed through a classification model that produces labeled events with a start time, end time, and a label. Moving forward, we shall refer to the first pipeline as AVC-FillerNet (for ASR + VAD + Classifier), and the second as VC-FillerNet (no ASR).
%
By skipping the ASR, VC-FillerNet is computationally lighter, but runs the risk of detecting parts of actual words as fillers \cite{das2019increase}.

Our goal is to train a robust multi-class classifier to detect fillers given a short snippet of audio. Given the label distribution in PodcastFillers, we opted to discard labels with 3K or less annotations and consolidate other labels, producing five new labels, each with ample training data: `filler' (`uh'+`um'), `words' (`regular words' and `repetitions'), `laughter', `music', and `breath'. Ultimately, we only care about the detection accuracy for the `filler' class, and we expect this consolidation to lead to a more robust classifier. In Section 5 we also evaluate our ability to classify `uh' and `um' as two separate classes.

We use wav2vec~\cite{schneider19wav2vec} embeddings computed with a 10 ms hop size as input to the model. Wav2vec was pretrained on over 960 hours of speech, providing a robust representation for classification of speech-like sounds. 
During training we apply time and ``frequency'' masking to the embeddings via SpecAugment~\cite{park19specaugment}, and optimize a cross entropy loss.


Since our goal is to detect specific short utterances in an audio stream, the task can be viewed as a keyword-spotting (KWS) problem where our keyword is the joint set of `uh' and `um'.
With this in mind,
we adapt a lightweight KWS model backbone architecture, TC-ResNet8~\cite{choi2019temporal}, for efficient classification. TC-ResNet8 only has around 100k parameters, making it suitable for low-latency inference. It applies 1D convolutions along the temporal axis and spans the entire frequency range in every layer, achieving strong performance even with a small number of layers. 
%
Because the filler candidates in AVC-FillerNet are normally short segments, we can train an \textit{event classifier} to directly predict the event label for the entire input segment. On the contrary, for VC-FillerNet, the filler candidates are usually long sequences of voice, so we train a \textit{frame classifier} to predict frame-level labels at a fine temporal resolution, e.g., every 100 ms.
To get frame-level predictions, we adapt the TC-ResNet8 backbone by adding an LSTM layer.
Similar to Filler-CRNN~\cite{das2019increase}, we then group contiguous frames with the same predicted label into an event. 
The final output of both the event-level classifier and frame-level classifier are discrete events with a start time, end time, and a label. We compare the two approaches as part of our ablation study.




\section{Experimental Design}

\label{sec:exps}
\subsection{Data split and training}
 For our experiments, we split the PodcastFillers dataset into train, validation, and test sets with 173, 6, 20 episodes respectively, while ensuring each subset remains gender-balanced.
 The audio is downsampled from 44.1 kHz to 16 kHz for computational efficiency.
 We train our proposed models on the training set, tune hyper-parameters and the VAD threshold on the validation set, and report performance on the test set. To train the event classifier, AVC-FillerNet, we use 1 s input clips with the labeled filler candidate placed at the center of the clip. The model produces a single prediction for the input, which is compared to the ground truth label. To train the frame classifier, VC-FillerNet, we use 1 s input clips where the filler candidate can appear anywhere in the clip. The model produces per-frame (10) predictions that are compared to the ground truth events (which have start/end times) frame-by-frame during training.

\subsection{Baselines}\label{sec:exp_baselines}
We compare our systems with two strong baselines: a neural-network-based method, Filler-CRNN~\cite{das2019increase}, and a forced-aligner-based method, Gentle~\cite{gentle}. The input to Filler-CRNN is log-mel with 128 bins computed from 1 s clips. 
The original Filler-CRNN architecture yielded weak performance in our experiments, so we fine tune it (number of layers, kernel size, pooling) on PodcastFillers for a stronger baseline. 
With Gentle, we first apply pre-trained Kaldi~\cite{povey2011kaldi} acoustic models developed on the Fisher English corpus~\cite{cieri2004fisher} to generate syllable tokens. We compare the tokens with the ASR transcript: filler words are detected if the inconsistent regions between the two are re-aligned by inserting `um' or `uh' into the transcript.

\subsection{Evaluation metrics}
\label{sec.evalm}
We compute segment-based and event-based metrics (Precision, Recall, F1) using sed$\_$eval~\cite{mesaros2016metrics} for the `filler' class, to evaluate detection accuracy and localization accuracy respectively.
Segment-based metrics map the system output and ground truth to a fixed time grid for comparison.
%
Event-based metrics compare the estimated sound events 
and the ground truth events directly. 
A predicted event is considered a true positive if it overlaps with a ground truth event that has the same label
\textit{and} its onset and offset are within a threshold (slack) from the reference event's onset and offset (200 ms in this work).

\subsection{Ablations} 
We run ablation studies to understand the impact of each of the two stages of our proposed pipeline, VAD and the filler classifier. For VAD, we vary the activation threshold from 0.1--0.9: the lower the threshold the more candidates will be passed to the second stage. For the classifier, we compare different input features (log-mel with 64 bins and wav2vec) and architectures (event classifier and frame classifier). We evaluate both the AVC-FillerNet and VC-FillerNet pipelines in all ablations.
%
%
The output events from each model are converted to frame-level likelihoods,
which are compared to the reference annotations to produce Precision-Recall (PR) curves,
which elucidate the trade-off between precision and recall. 

\section{Results} 

\subsection{Ablation studies}
\label{sec:ab}

\begin{figure}[t]
\setlength{\belowcaptionskip}{-8pt}
\hspace{-0.25cm}
\includegraphics[width=3in]{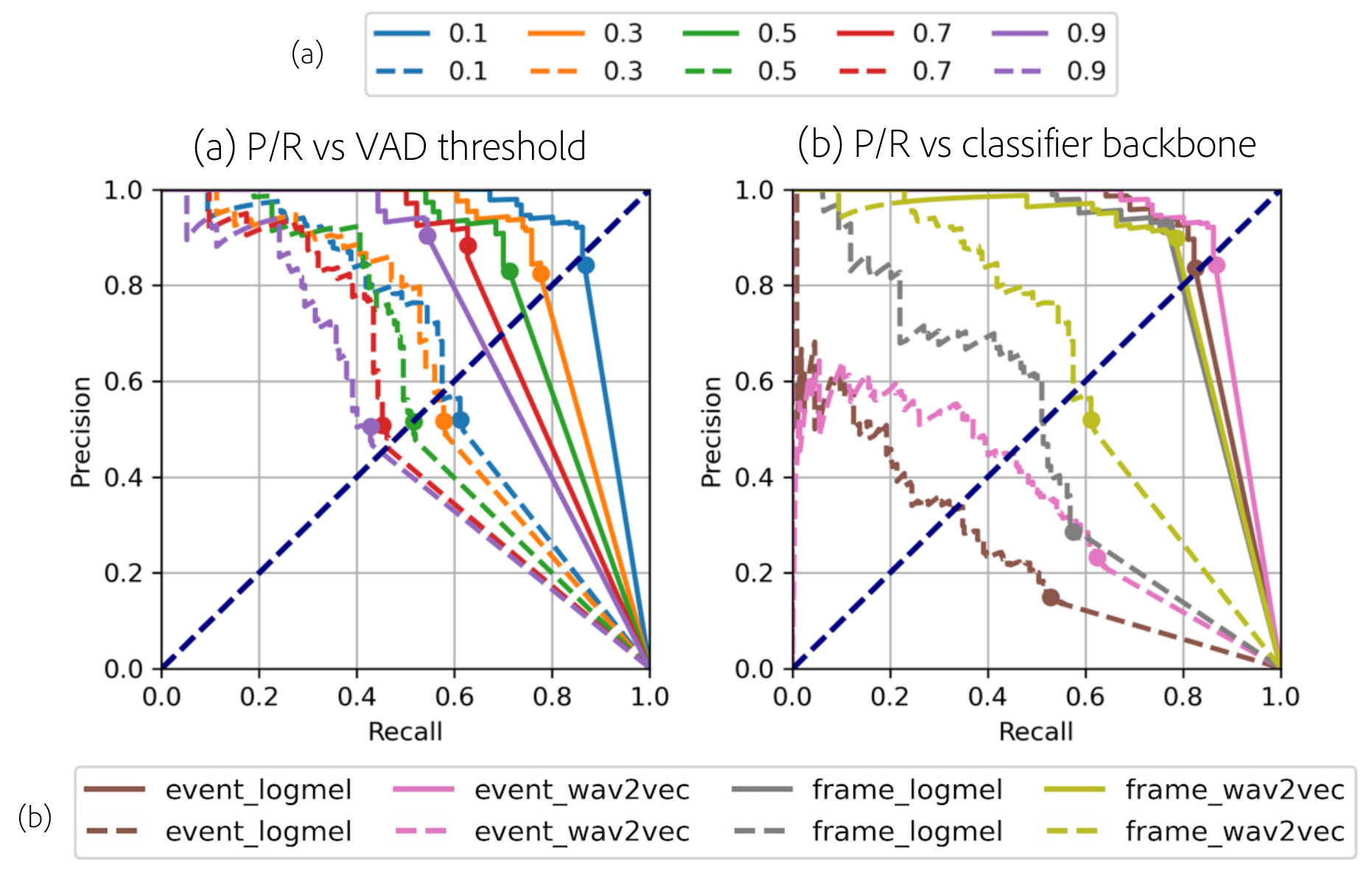}
\caption{Frame level P/R vs (a) VAD threshold and (b) classifier backbone. 
`event' stands for event classifier and `frame' for frame classifier, solid lines are AVC-FillerNet, dashed lines are VC-FillerNet. Dotted position is the classifier threshold at 0.5.}
\label{fig:abs}
\end{figure}

We start by analyzing the influence of the VAD activation threshold, using the PodcastFillers validation set, and wav2vec as the input feature. 
The results are shown in Fig.~\ref{fig:abs}(a). The best PR-curve is obtained using the lowest threshold (0.1), and as we increase the threshold there is a notable decrease in recall.
%
Interestingly, the precision remains consistent regardless of the VAD activation threshold. This suggests that the filler classifier is robust at rejecting false positives with lower VAD likelihoods, and so a low VAD threshold maximizes recall by ensuring we do not miss soft filler words, without compromising on precision.
%


Following this,
we fix the VAD threshold to 0.1 in the second ablation study where we compare different input features and classifier backbones, shown in Fig.~\ref{fig:abs}(b). 
Wav2vec consistently outperforms log-mel as the input feature, 
confirming that this model, trained on a large speech corpus, yields a discriminative representation for filler word classification.
For the AVC-FillerNet pipeline, the event classifier is only marginally better than the frame classifier. In contrast, since the VC-FillerNet pipeline cannot leverage ASR to determine the precise timing of filler candidates, the frame classifier outperforms the event classifier in this pipeline due to its superior temporal
accuracy.
%

Most importantly, we see that across both ablation studies, AVC-FillerNet clearly outperforms VC-FillerNet. By leveraging ASR, AVC-FillerNet produces tight temporal boundaries around filler candidates, 
and  dramatically reduces the number of candidates passed to the classifier. 
This reduces the chances of producing false positives, leading to a boost in precision.



\subsection{Comparison to baselines}
We compare AVC-FillerNet and VC-FillerNet with two baselines, Filler-CRNN and Gentle (described in Sec~\ref{sec:exp_baselines}), and report results the PodcastFillers test set as shown in Tab.~\ref{tab:syscomp}. 
For our pipelines, we use the optimal VAD threshold of 0.1 as determined by the ablation study on the validation set.
We see that AVC-FillerNet significantly outperforms all the other systems for all metrics. Gentle yields higher precision than VC-FillerNet and Filler-CRNN, but has the lowest recall among all the systems. 
We speculate this may be improved by leveraging acoustic models trained with filler words in Gentle.



\begin{table}[t]
\caption{
Segment- and event-based evaluation results ($\%$) for our proposed systems and baselines for filler word detection.
} 
\setlength{\tabcolsep}{3.5pt}
\centering
\begin{tabular}{c|cccccc}
\toprule
\multirow{2}{*}{\textbf{System}} & \multicolumn{3}{c}{\textbf{Segment}} & \multicolumn{3}{c}{\textbf{Event}}  \\
\cline{2-7}
&P&R&F1&P&R&F1\\
\hline
\textbf{AVC-FillerNet (Ours)}   &\textbf{93.0}&\textbf{95.4}&\textbf{94.2}&\textbf{91.7}&	\textbf{94.0}& \textbf{92.8}	 \\ 
\textbf{VC-FillerNet (Ours)}&71.6&71.0 &71.3&66.0&76.9&71.0	 \\ 
Filler-CRNN~\cite{das2019increase}   &56.4& 70.3&62.6&37.5 &78.3&50.7	\\ 
Gentle~\cite{gentle}   &78.4&64.8&71.0&77.0&64.9&70.4 \\ 
 \bottomrule
\end{tabular}
\label{tab:syscomp}
\end{table}

\subsection{Filler detection \& classification with fine granularity}
Finally, we evaluate our systems' ability to separately detect `uh' and `um' filler words. We re-train the classifier with the two labels as separate classes, and evaluate our systems on the PodcastFillers test set, as shown in Tab.~\ref{tab:filler}. We see that `um' is easier to classify, especially for VC-FillerNet. We speculate that this is because `um's are typically longer than `uh's, providing the classifier with more signal to leverage for inference.

\begin{table}[h]
\caption{Segment- and event-based F1 measure ($\%$) results for separately detecting `uh' and `um' with our proposed systems.} 
\setlength{\tabcolsep}{3.5pt}
\centering
\begin{tabular}{c|cccc}
\toprule
\multirow{2}{*}{\textbf{System}} & \multicolumn{2}{c}{\textbf{`Um'}} & \multicolumn{2}{c}{\textbf{`Uh'}}  \\
\cline{2-5}
&Segment&Event&Segment&Event\\
\hline
\textbf{AVC-FillerNet}  &92.5 &91.0&85.0& 84.3	 \\ 
\textbf{VC-FillerNet} &75.2&75.9&57.0&57.1	 \\ 
 \bottomrule
\end{tabular}
\label{tab:filler}
\end{table}

\section{Conclusion} 
\label{sec:conclusion}
In this work we presented PodcastFillers, a large dataset of podcasts with annotated filler words. 
The dataset was created by boostrapping a VAD model and a commercial ASR system to generate filler candidates that were annotated via crowdsourcing. We proposed ASR-based and ASR-free filler detection and classification pipelines, AVC-FillerNet and VC-FillerNet. Our experiments showed that AVC-FillerNet achieves state-of-the-art results, significantly outperforming existing filler word detection systems, and that leveraging ASR outperforms a keyword spotting approach for filler word detection. Through ablation studies, we evaluated the impact of our design choices on system performance. We hope the PodcastFillers dataset, our proposed filler detection and classification pipeline, and our experimental results serve as a benchmark for future research.


\bibliographystyle{IEEEtran}
\bibliography{mybib}

\onecolumn

\begin{appendices}

We provide additional information about the annotation interface we built to label PodcastFillers (Appx.~\ref{sec:ann}), the distribution of voice pitch in the dataset (Appx.~\ref{sec:pitch}), and the model architectures used for our VAD and filler classification models (Appx.~\ref{sec:backbones}). In Appx.~\ref{sec:confusion} we present confusion matrices produced by the filler classification model, to provide further insight into the model's performance.

\section{Annotation interface}
\label{sec:ann}
\begin{figure}[h]
\centering
\setlength{\belowcaptionskip}{-8pt}
\hspace{-0.25cm}
\includegraphics[width=6.1in]{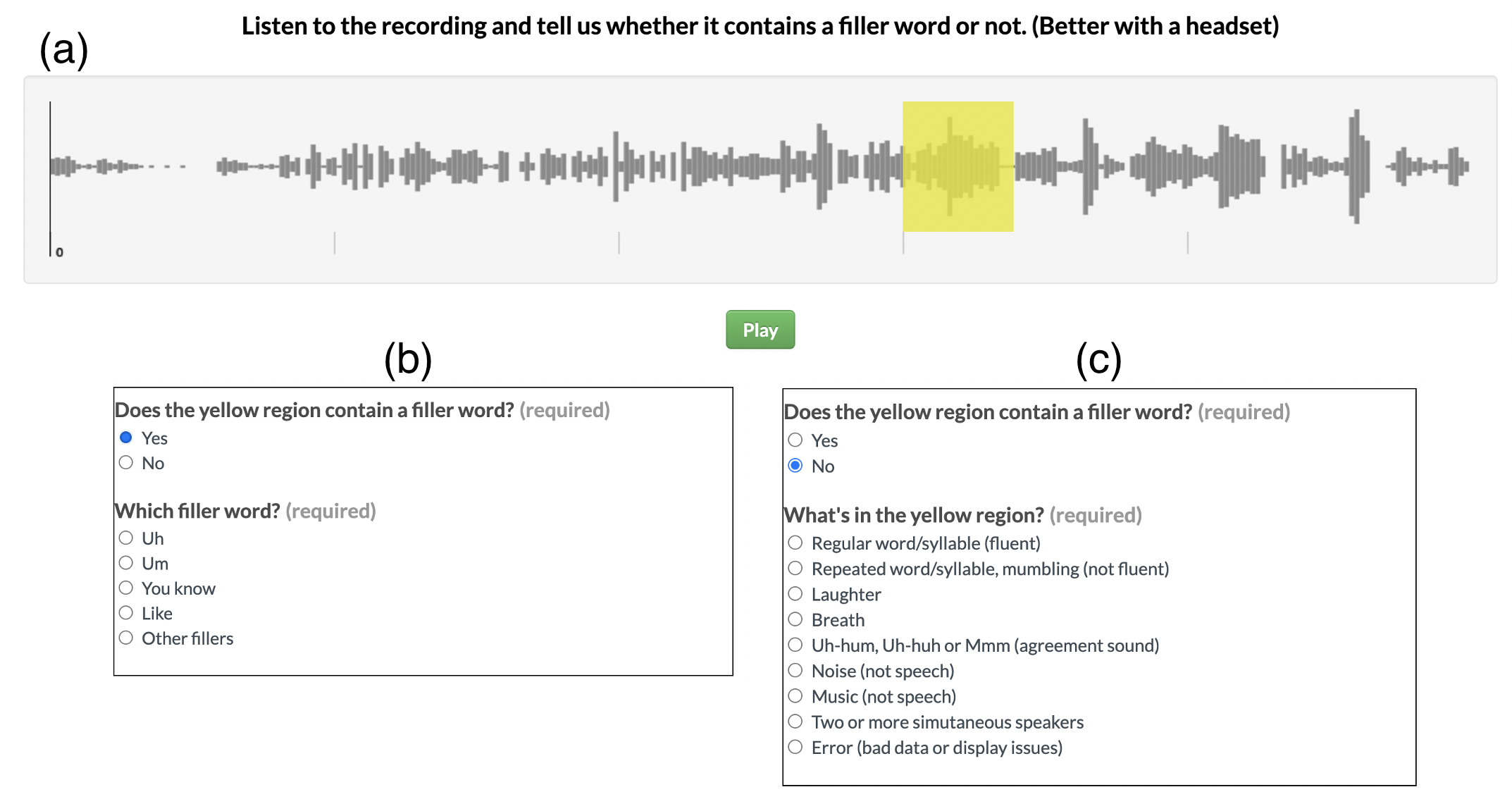}
\vspace{-5pt}
\caption{Our custom-built filler word annotation interface used to annotate the PodcastFillers dataset. (a) Audio player with a filler candidate at time 3 s highlighted in yellow, (b) fine-grained filler labels, (c) fine-grained non-filler labels.}
\label{fig:int}
\end{figure}

To annotate the PodcastFillers dataset, we custom-built the annotation interface depicted in Fig.~\ref{fig:int}. Annotators were presented with an audio player with a waveform visualization of a 5 s audio clip, with a filler word candidate at time 3 s highlighted in yellow (Fig.~\ref{fig:int}(a)). Annotators were first asked if the yellow region contains a filler word or not. Based on their answer, a second list of options appeared for them to label the fine-grained filler or non-filler type, depicted in Fig.~\ref{fig:int}(b) and Fig.~\ref{fig:int}(c) respectively.

\vspace{-5pt}

\section{Pitch distribution}
\label{sec:pitch}
\begin{figure}[t]
\hspace{-0.25cm}
\begin{subfigure}{0.25\textwidth}
\centering
\includegraphics[width=1.0\textwidth]{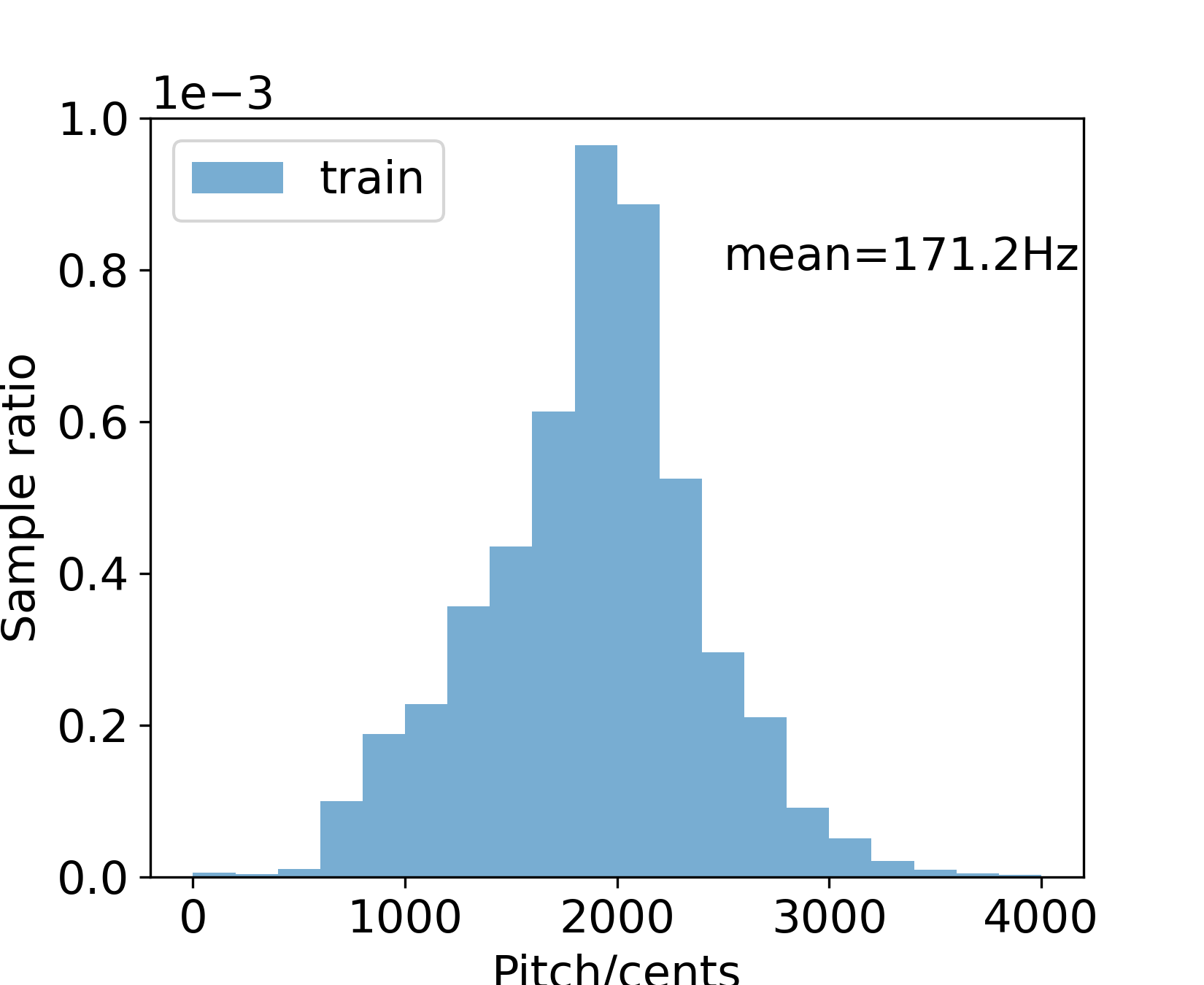}
\caption{Training set}
\label{fig:pd_train}
\end{subfigure}\hfill
\begin{subfigure}{0.25\textwidth}
\centering
\includegraphics[width=1.0\textwidth]{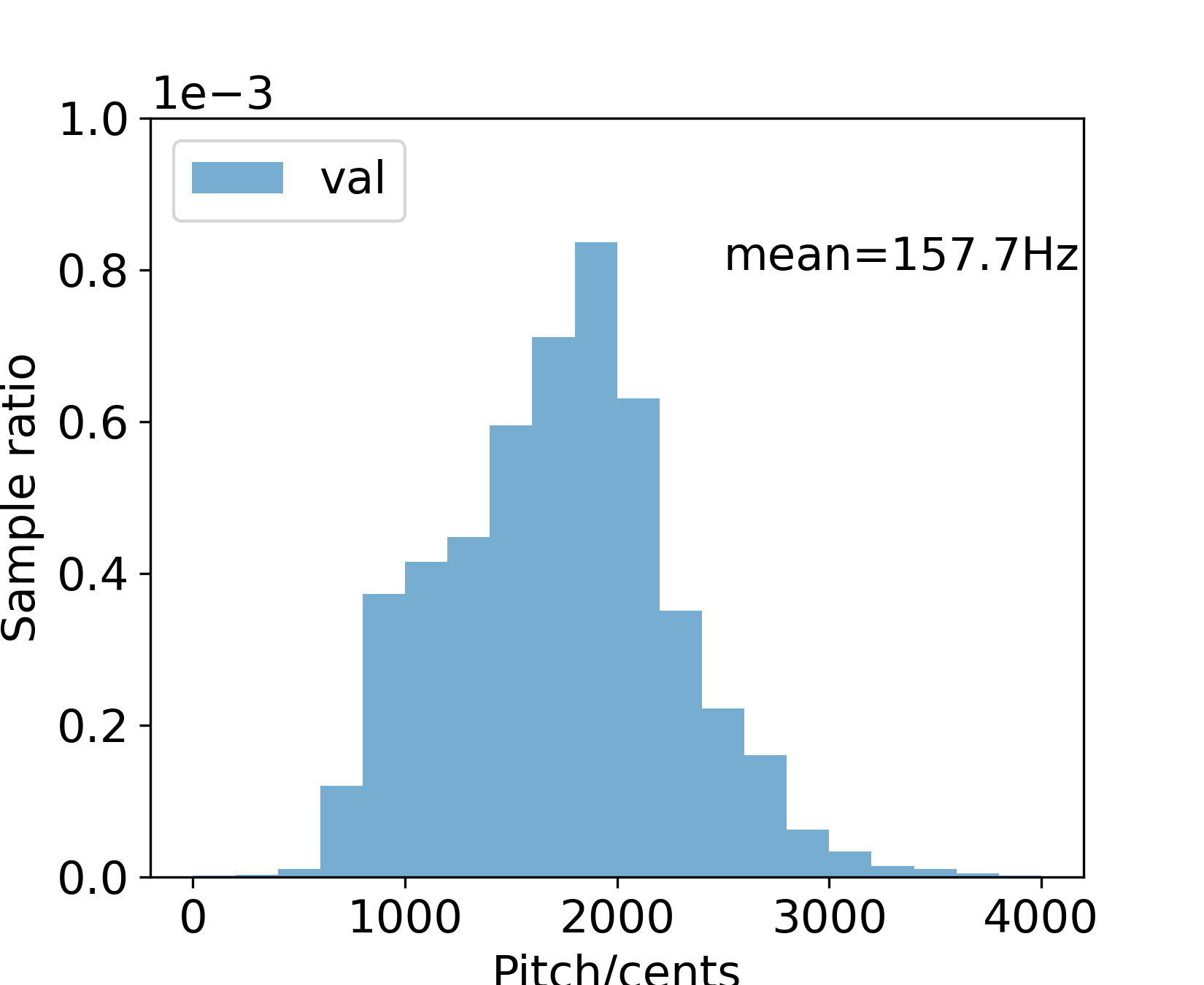}
\caption{Validation set}
\label{fig:pd_val}
\end{subfigure}\hfill
\begin{subfigure}{0.25\textwidth}
\centering
\includegraphics[width=1.0\textwidth]{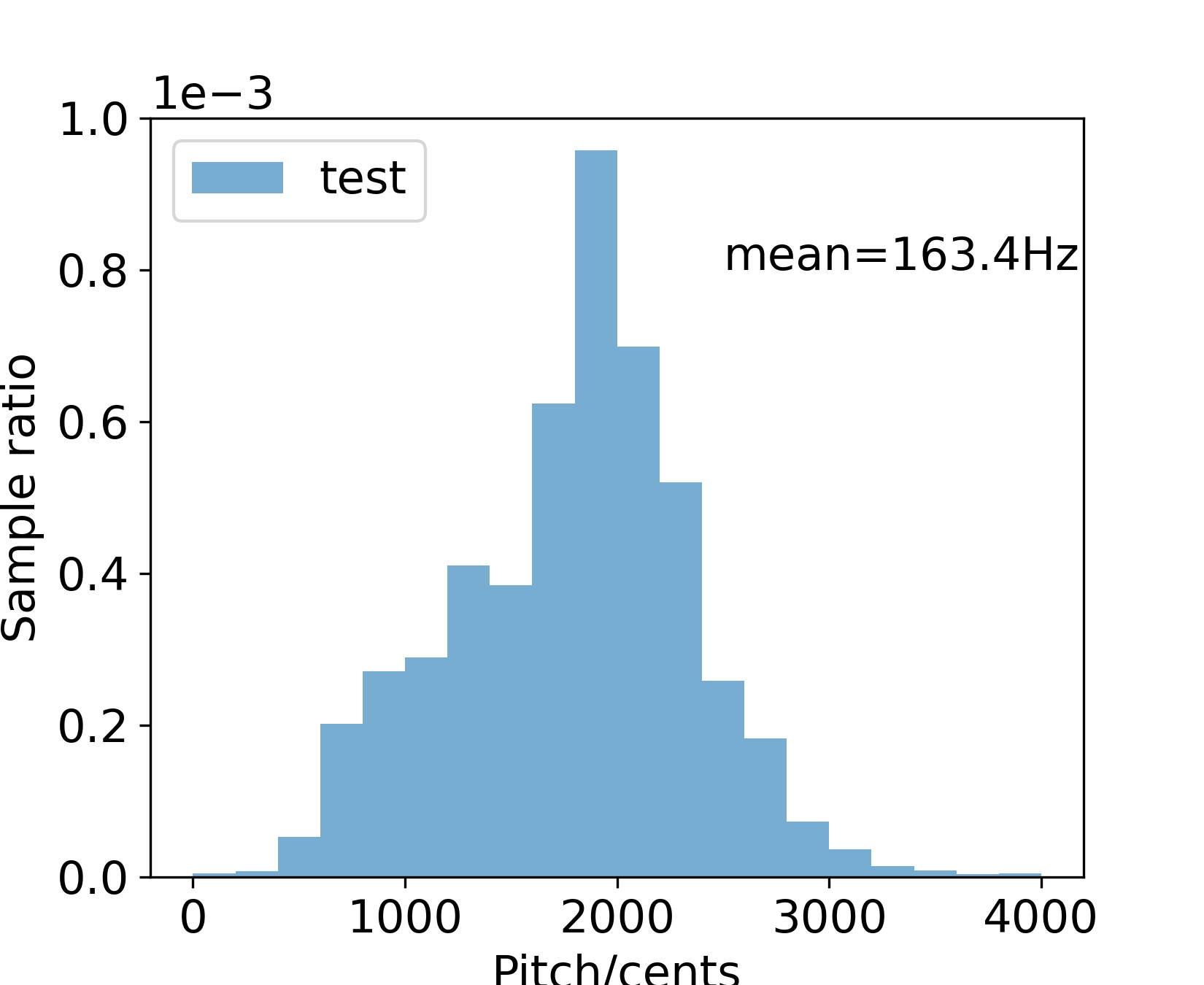}
\caption{Test set}
\label{fig:pd_test}
\end{subfigure}\hfill
\begin{subfigure}{0.25\textwidth}
\centering
\includegraphics[width=1.0\textwidth]{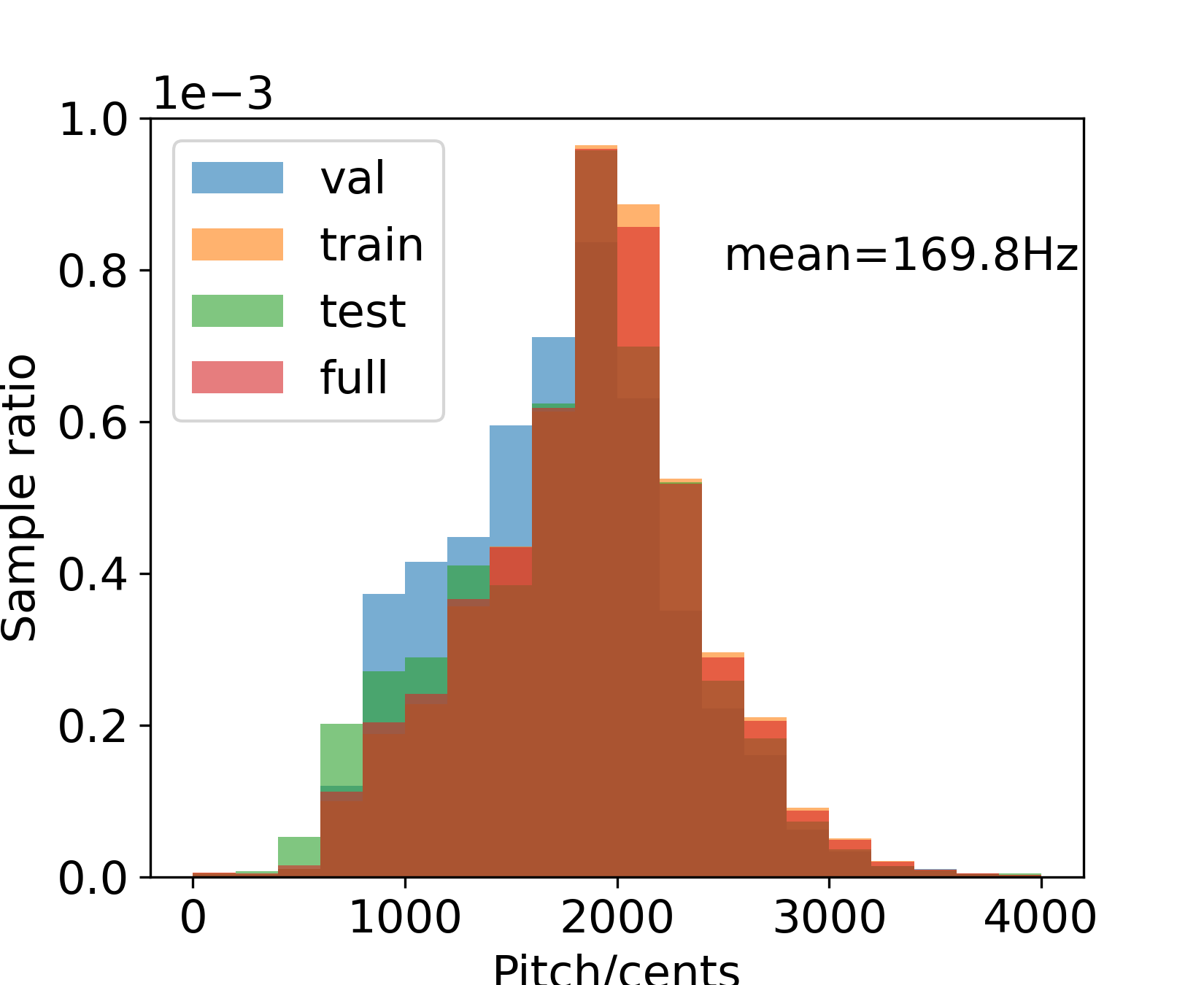}
\caption{Entire dataset}
\label{fig:pd_test}
\end{subfigure}
\caption{Pitch distribution of filler candidates in the (a) train, (b) validation, (c) test sets of PodcastFillers, and (d) the entire dataset.}
\label{fig:pd}
\end{figure}

When curating PodcastFillers dataset, we ensured the pitch ($f_0$) of the voices in the dataset spans the range of adult speech pitch from 60 Hz to 300 Hz. We used CREPE\footnote{J.W.~Kim, J.~Salamon, P.~Li, and J.P.~Bello, ``CREPE: A convolutional
representation for pitch estimation,'' in ICASSP, 2018.}
for pitch estimation, and measure pitch in cents relative to 55 Hz. The pitch distribution for all filler candidates in the dataset is depicted in Fig.~\ref{fig:pd}(d), with a mode roughly at the center of the pitch range, around 170 Hz. The dataset contains roughly equally decreasing numbers of samples as we move away from the mode toward higher and lower voices. The train, validation and test splits, depicted in subplots (a), (b), (c) respectively, exhibit a similar pitch distribution for filler candidates.

\section{Model architectures for VAD and the filler word classifier}
\label{sec:backbones}
\begin{figure}[t]
\hspace{-0.25cm}
\centering
\begin{subfigure}{0.5\textwidth}
\centering
\includegraphics[width=1.6in]{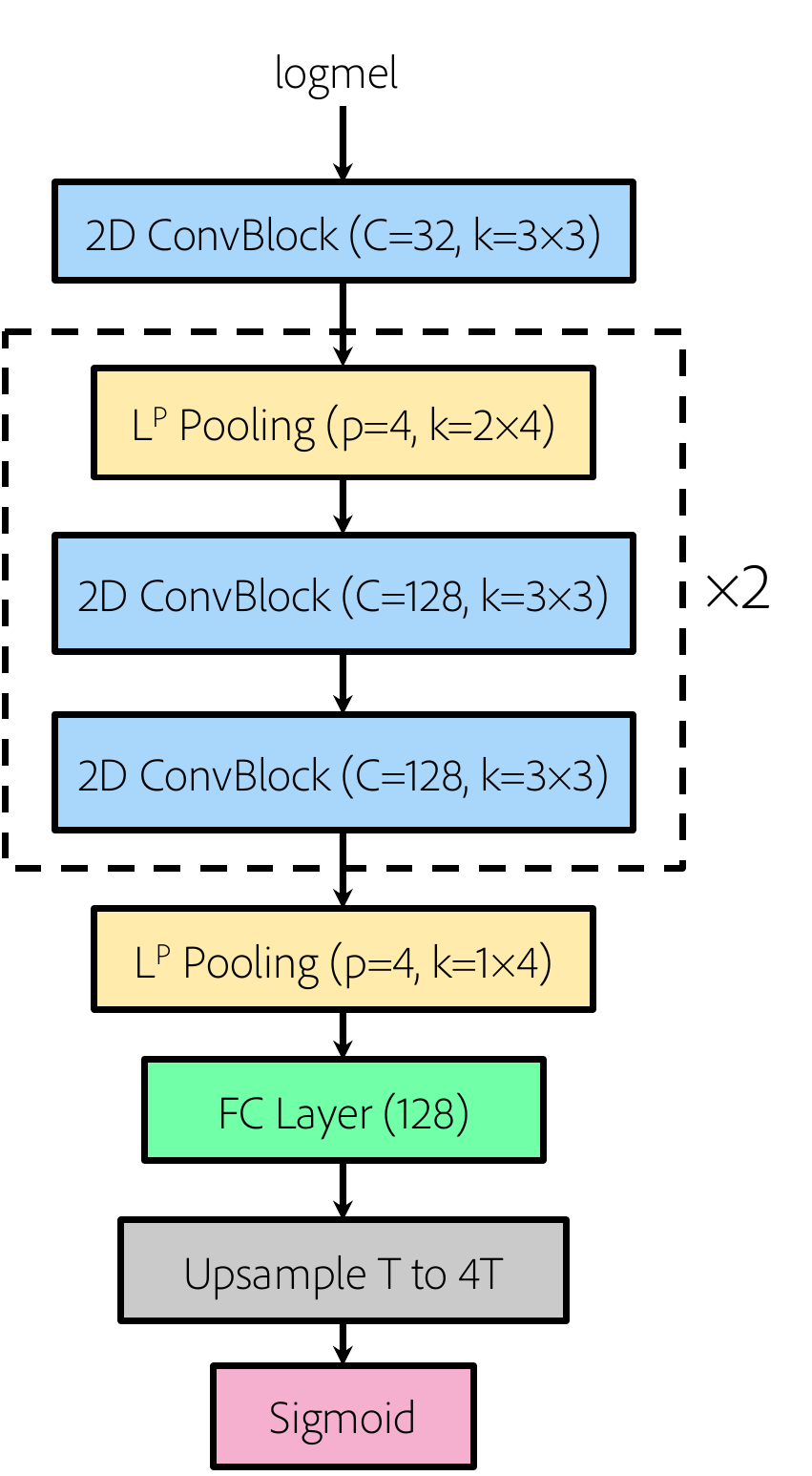}
\label{fig:vad}
\caption{}
\end{subfigure}\hfill
\begin{subfigure}{0.5\textwidth}
\centering
\includegraphics[width=2.4in]{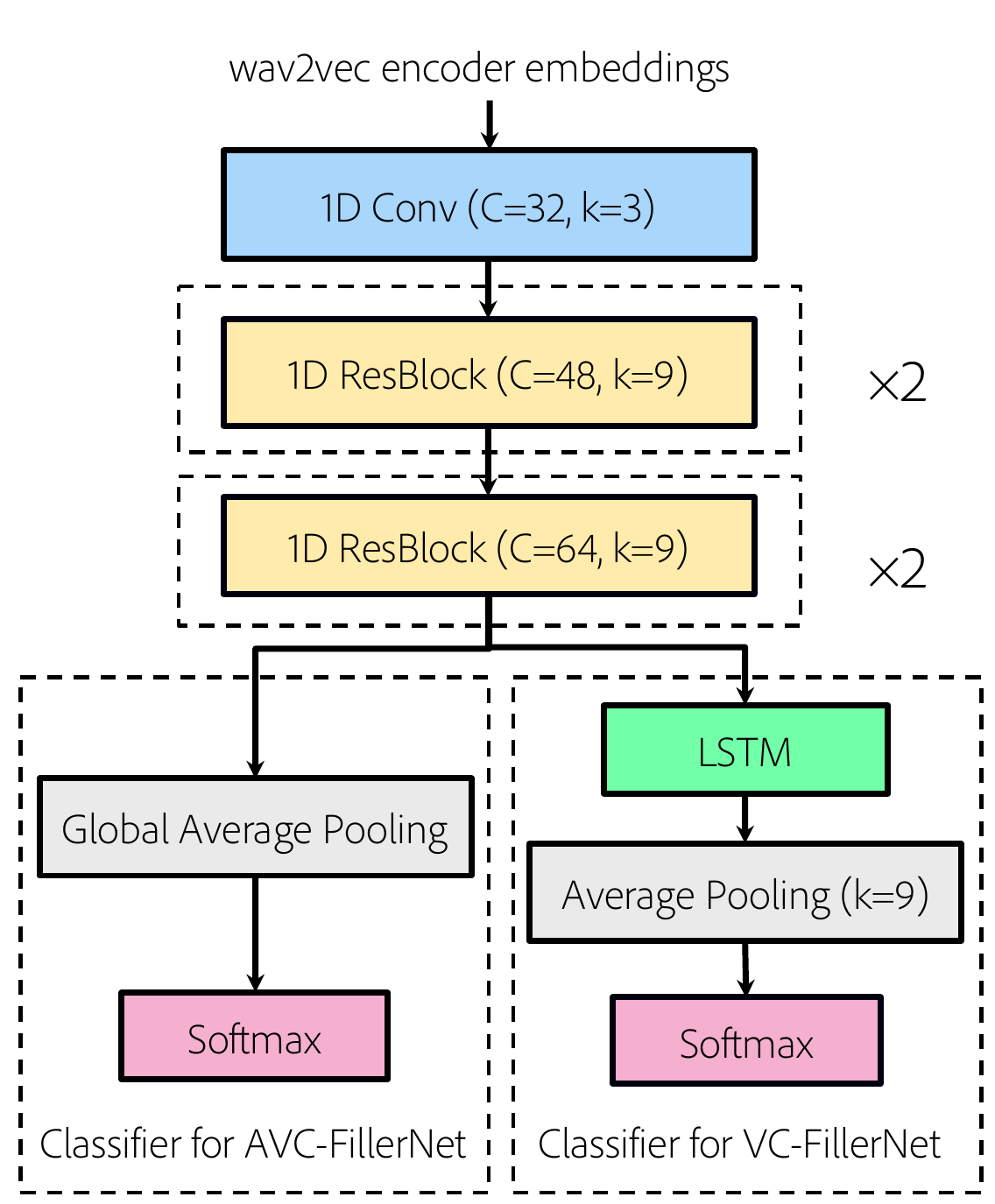}
\label{fig:fc}
\caption{}
\end{subfigure}
\caption{(a) Voice activity detection (VAD) network architecture used in both AVC-FillerNet and VC-FillerNet. (b) filler classifier network architecture used for AVC-FillerNet (left fork) and VC-FillerNet (right fork).}
\label{fig:backbones}
\end{figure}

The network architecture for the VAD model described in Section \ref{sec:data}, adapted from~\cite{Chen2020voice}, is depicted in Fig.~\ref{fig:backbones}(a). The two variants of the architecture used for the filler classifier described in Section \ref{sec:method}, a TCResNet-8~\cite{choi2019temporal}, are depicted in Fig.~\ref{fig:backbones}(b).

\section{Confusion matrices for the filler word classifier}
\label{sec:confusion}
\begin{figure}[b]
\hspace{-0.25cm}
\begin{subfigure}{0.25\textwidth}
\centering
\includegraphics[width=1.0\textwidth]{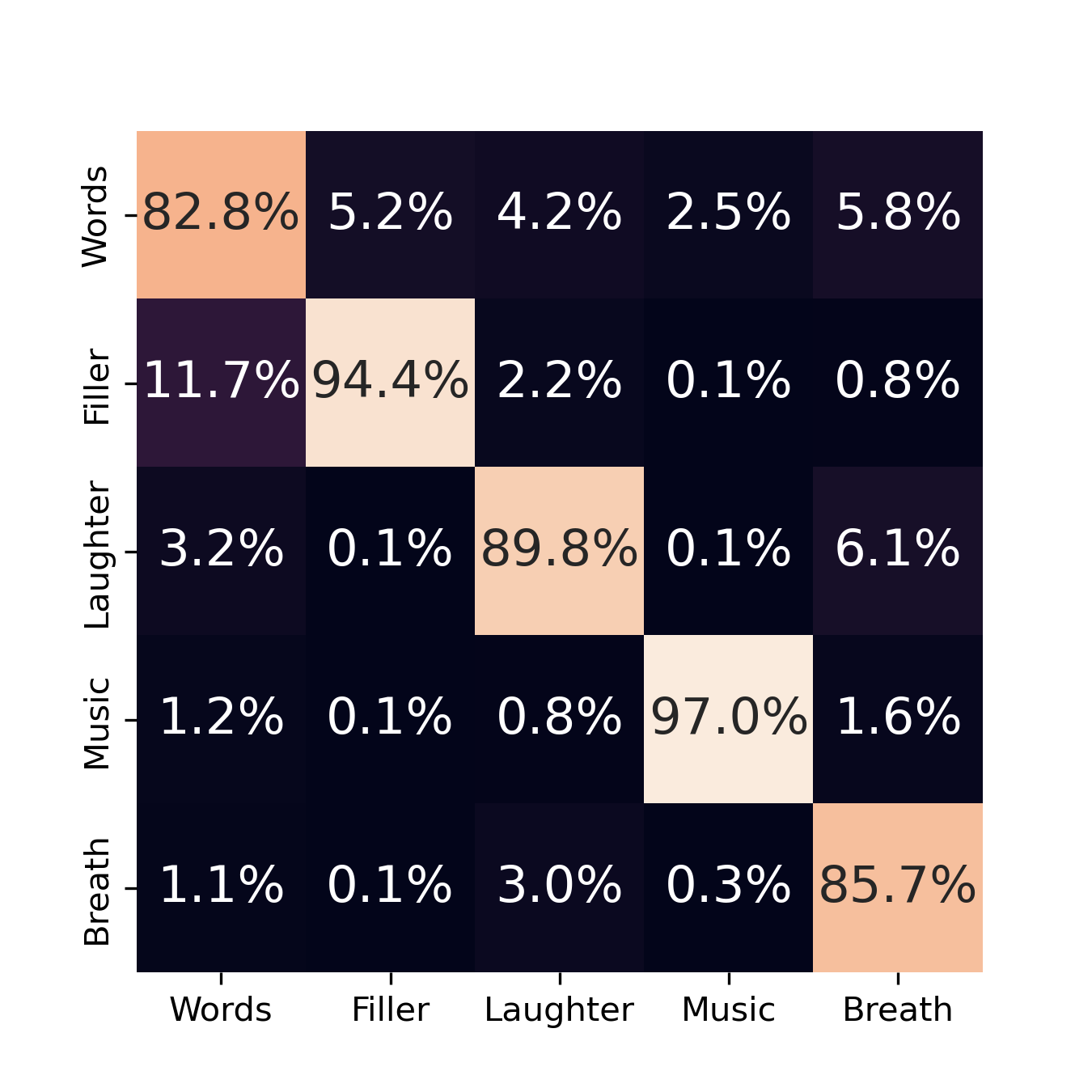}
\caption{wav2vec, coarse labels}
\label{fig:cm_5_wav2vec}
\end{subfigure}\hfill
\begin{subfigure}{0.25\textwidth}
\centering
\includegraphics[width=1.0\textwidth]{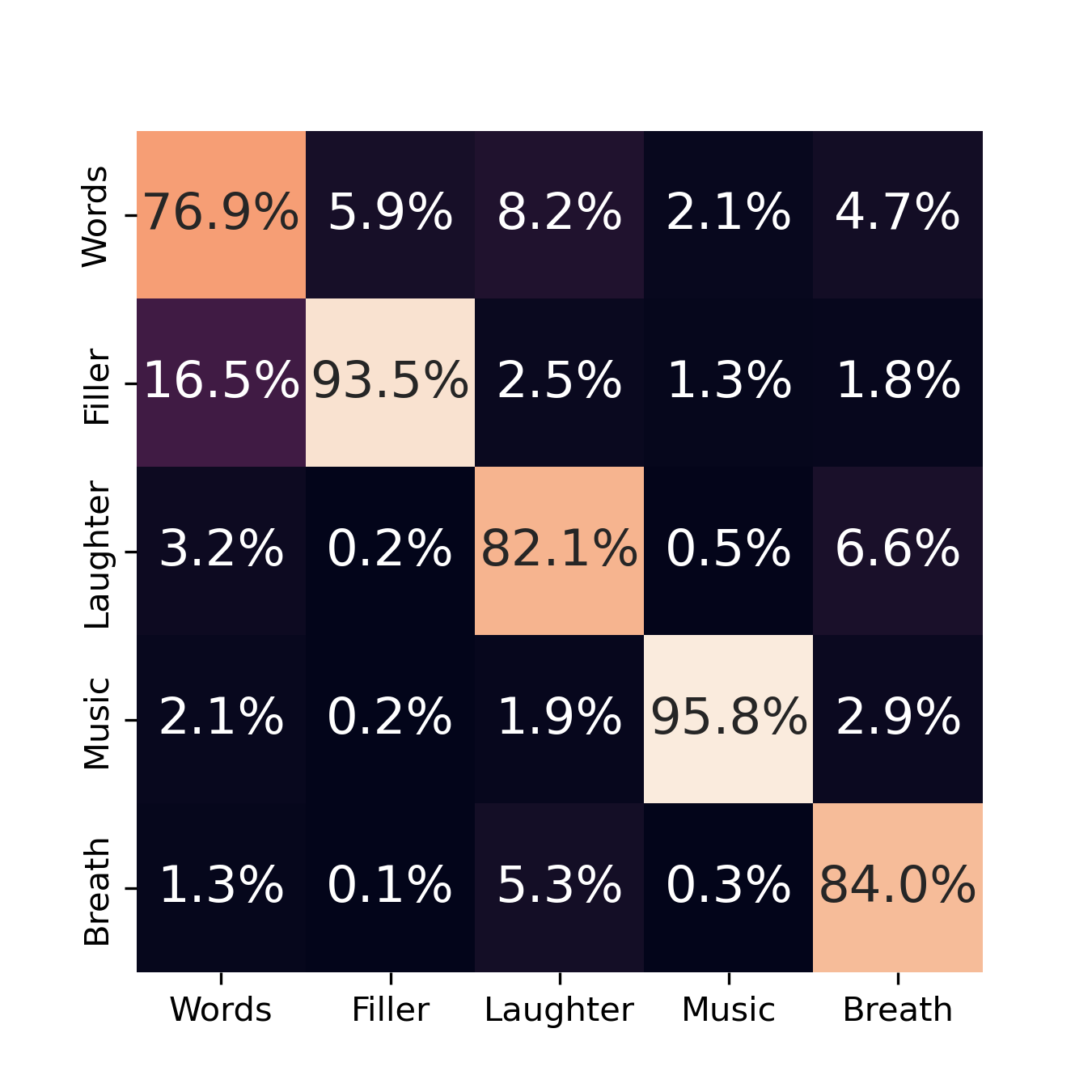}
\caption{log-mel, coarse labels}
\label{fig:cm_5_logmel}
\end{subfigure}\hfill
\begin{subfigure}{0.25\textwidth}
\centering
\includegraphics[width=1.0\textwidth]{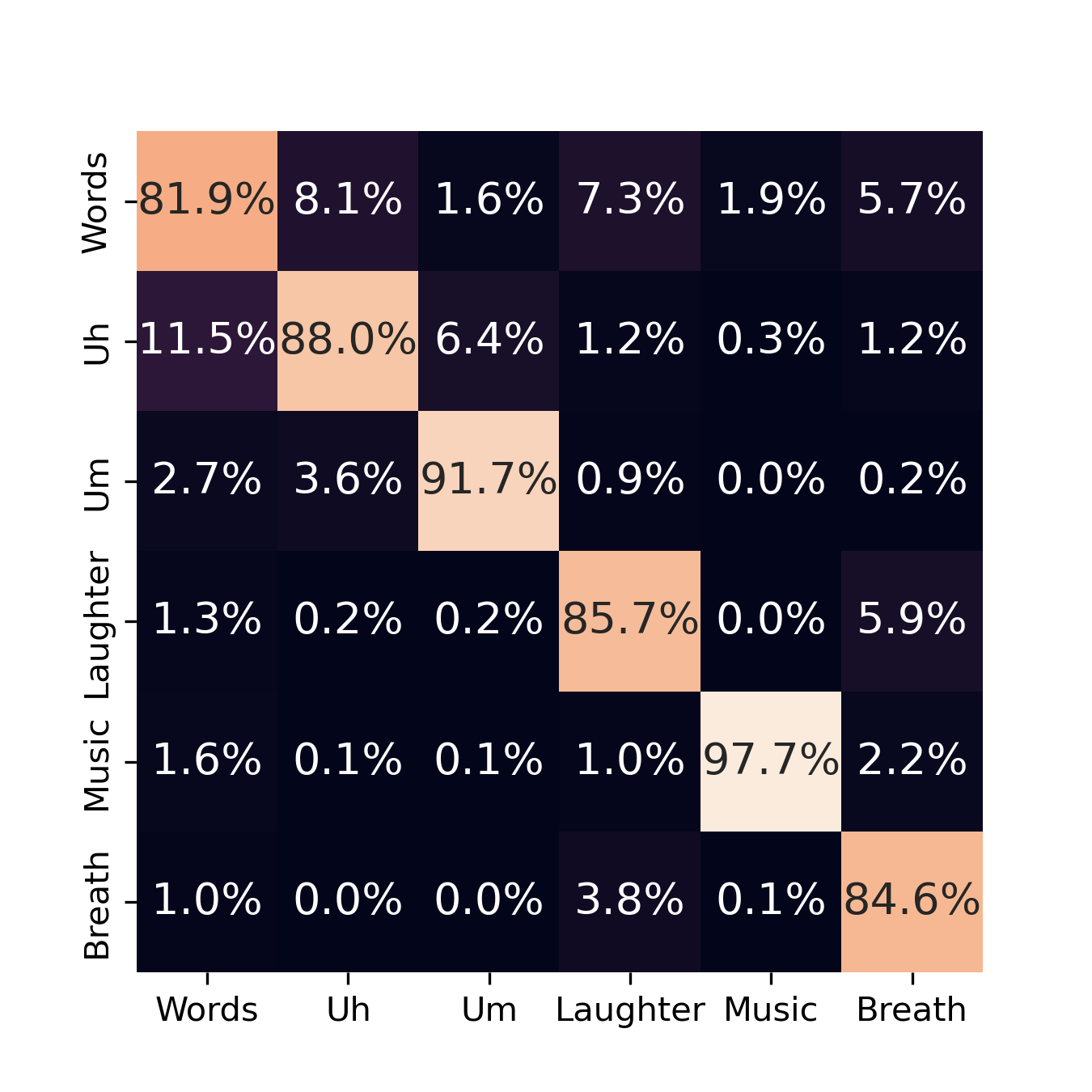}
\caption{wav2vec, granular labels}
\label{fig:cm_6_wav2vec}
\end{subfigure}\hfill
\begin{subfigure}{0.25\textwidth}
\centering
\includegraphics[width=1.2\textwidth]{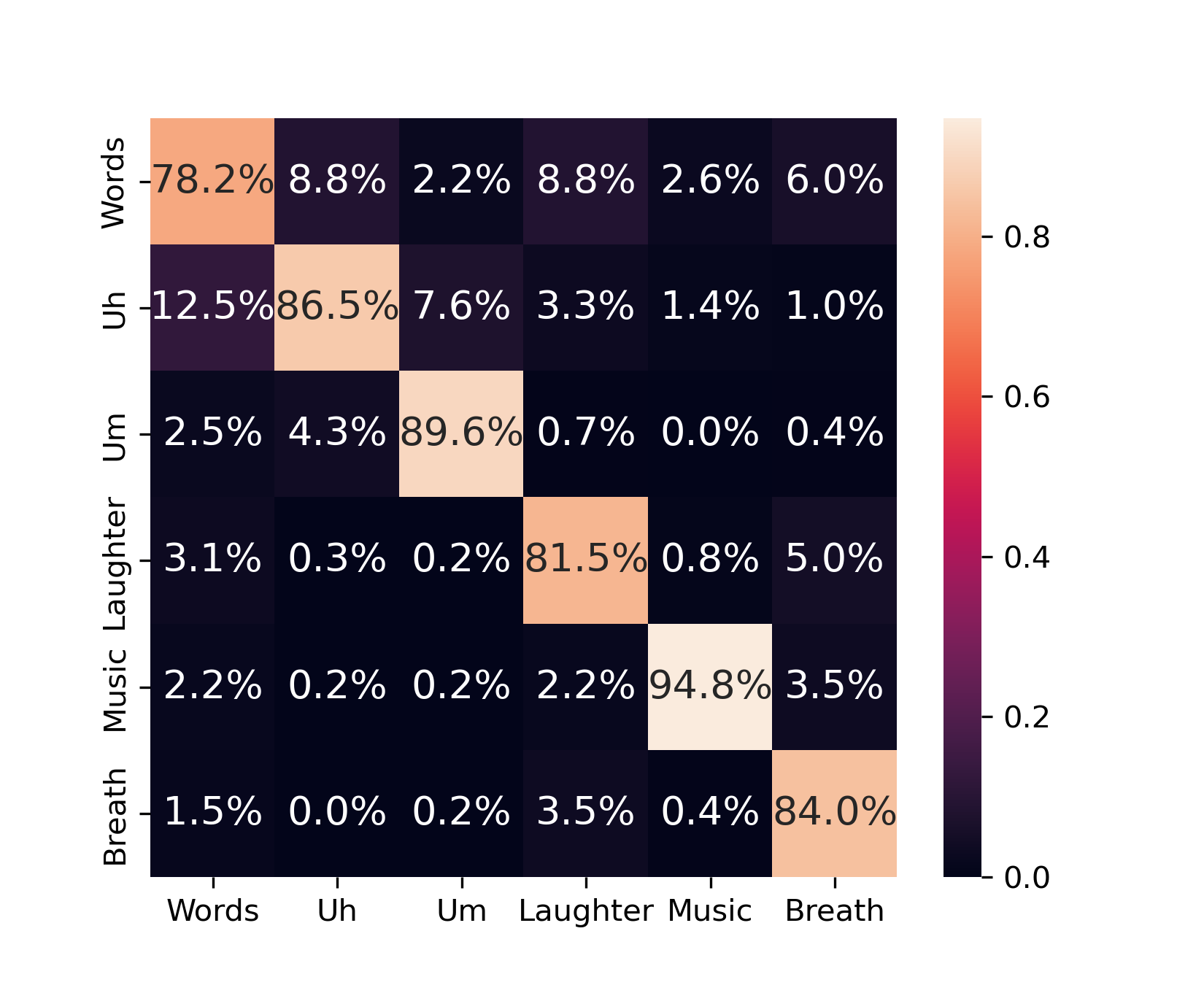}
\caption{log-mel, granular labels}
\label{fig:cm_6_logmel}
\end{subfigure}
\caption{Confusion matrices for the filler classifier used in AVC-FillerNet for different input features and target labels.}
\label{fig:confusionmatrices}
\end{figure}

In Fig.~\ref{fig:confusionmatrices} we depict the confusion matrices produced by the filler classification model used in the AVC-FillerNet pipeline when evaluated on 1 s filler candidate clips from the PodcastFillers test set. We use either wav2vec or log-mel as the input feature to the model, and predict either a coarse label set with 5 labels where `uh' and `um' are grouped into a `filler` label, or a granular label set with 6 classes where `uh' and `um' are treated as separate labels.
For the coarse labels, the greatest source of confusion is between `filler' and `words', which makes sense given their acoustic similarity. For the granular labels, `uh' is confused with words more often than `um' is. We conjecture the additional `m' in the latter makes it easier to classify. For both label sets, using wav2vec as the input feature reduces the confusion between filler words and regular words compared to using log-mel as the input feature.

\end{appendices}
\end{document}